\pdfoutput=1

\documentclass[11pt]{article}

\usepackage[]{EMNLP2023}

\usepackage{times}
\usepackage{latexsym}

\usepackage[T1]{fontenc}

\usepackage[utf8]{inputenc}

\usepackage{microtype}

\usepackage{inconsolata}

\usepackage{graphicx}
\usepackage{subcaption}
\usepackage{dblfloatfix} 
\usepackage{multirow} 

\usepackage{makecell}

\usepackage{algorithm}
\usepackage{listings}
\usepackage{xcolor}

\definecolor{codegray}{rgb}{0.5,0.5,0.5}
\definecolor{codepurple}{rgb}{0.58,0,0.82}
\definecolor{codegreen}{rgb}{0,0.58,0.22}

\lstdefinestyle{mystyle}{
    commentstyle=\color{codegray},
    keywordstyle = \color{blue},
    numberstyle=\tiny\color{codegray},
    stringstyle=\color{codegreen},
    basicstyle=\ttfamily\footnotesize,
    breakatwhitespace=false,         
    breaklines=true,                 
    captionpos=b,                    
    keepspaces=true,                 
    numbers=left,                    
    numbersep=5pt,                  
    showspaces=false,                
    showstringspaces=false,
    showtabs=false,                  
    tabsize=1
}

\usepackage{booktabs}

\title{Pose-Based Sign Language Appearance Transfer}

\author{Amit Moryossef\textsuperscript{$1,2\ast$}, Gerard Sant\textsuperscript{$1\ast$}, Zifan Jiang\textsuperscript{1} \\
  \textsuperscript{1}University of Zurich, \textsuperscript{2}\href{http://sign.mt}{sign.mt} \\
  \texttt{amit@sign.mt}}

\begin{document}
\maketitle


\begin{abstract}
We introduce a method for transferring the signer's appearance in sign language skeletal poses while preserving the sign content. Using estimated poses, we transfer the appearance of one signer to another, maintaining natural movements and transitions. This approach improves pose-based rendering and sign stitching while obfuscating identity. Our experiments show that while the method reduces signer identification accuracy, it slightly harms sign recognition performance, highlighting a tradeoff between privacy and utility.
Our code is available at \url{https://github.com/sign-language-processing/pose-anonymization}.
\end{abstract}

\section{Introduction}

Personal data, particularly person-identifying information, is central to data protection laws in many countries, including the EU General Data Protection Regulation (GDPR; \citet{GDPR2016a}). In signed languages, identifying information is embedded in every utterance through appearance, prosody, movement patterns, and sign choices \citep{bragg2020, battisti-etal-2024-person}.
Therefore, from an information-theoretic perspective, removing all identifying information necessitates removing all information. However, a tradeoff between privacy and utility can be achieved by selectively removing some information.

\begin{figure}[!hb]
    \centering
    \includegraphics[width=0.7\linewidth]{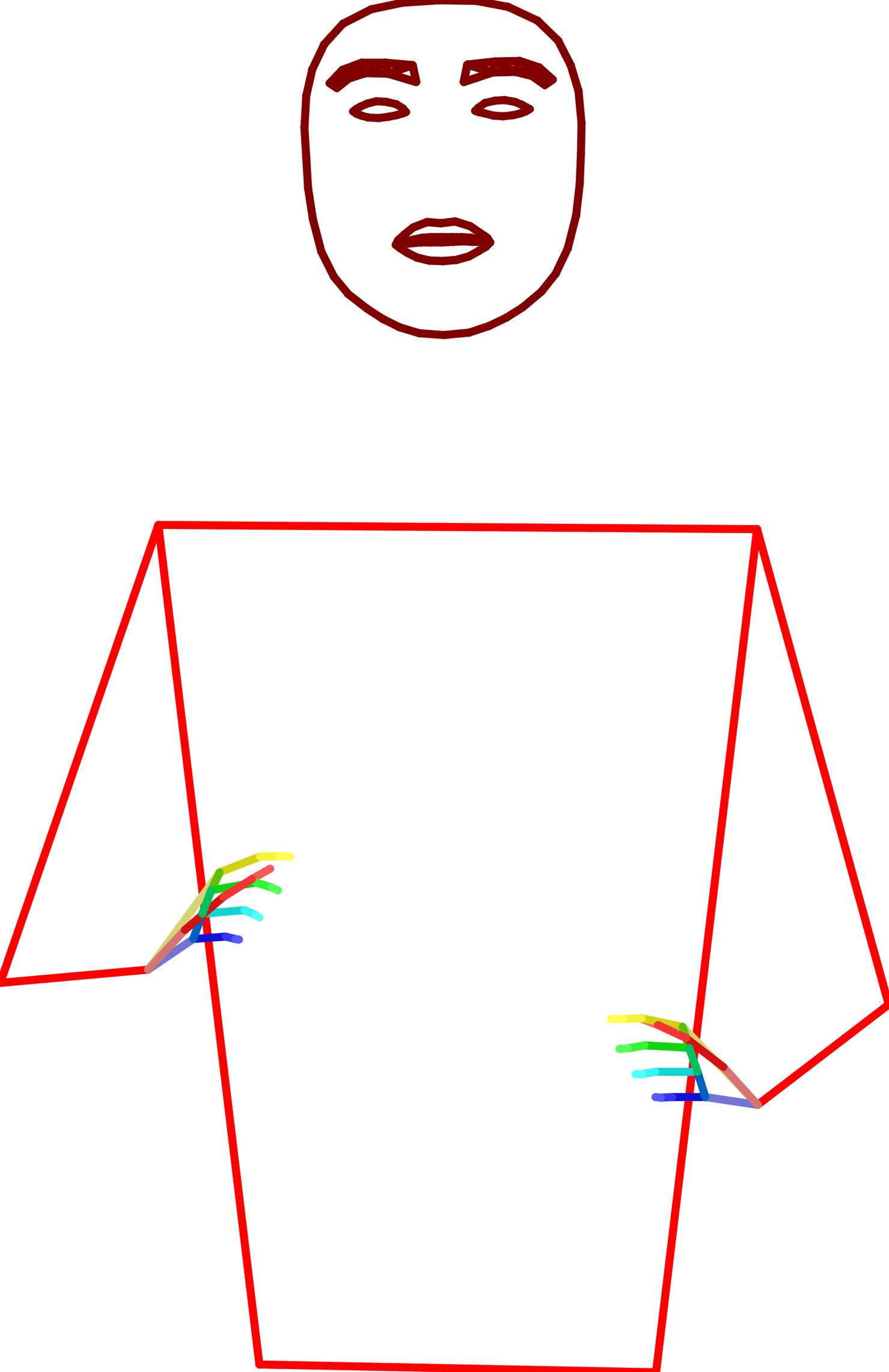}
    \caption{The average MediaPipe Holistic frame (landmarks reduced for visual clarity) extracted from a large sign language dataset ($\approx50$ million frames).}
    \label{fig:mean}
\end{figure}

We propose a straightforward yet effective method for altering the appearance of a signer in a sign language pose (Figure \ref{fig:mean}) while preserving the underlying sign content (\S\ref{sec:method}). Specifically, given a sign language video by signer $\alpha$ and an image of person $\beta$, our method generates the appearance of person $\beta$ performing the same signs as signer $\alpha$.  

Qualitatively, this method effectively smooths skeletal pose stitching \cite{moryossef2023baseline}, and improves pose-based video rendering \cite{anonysign}. However, quantitative evaluation of our method as data augmentation reveals that while it can help confuse signer identification models, it hurts sign language recognition (\S\ref{sec:experiments}).

\section{Related Work}\label{sec:background}

Research on sign language poses appearance varies in purpose. As \citet{Isard2020ApproachesTT} highlights, video anonymization falls into two main categories: concealing parts of the video \citep{dataset:hanke-etal-2020-extending, Rust2024TowardsPS} or reproducing the video without certain information. This work focuses on the latter.

For instance, \citet{anonysign} replace the signer's visual appearance, targeting human consumption. They estimate poses from the original video and use a Generative Adversarial Network (GAN; \citet{goodfellow2014generative}) to generate a different-looking human. This process, working correctly, anonymizes the signing video as effectively as pose estimation alone, since all of the information from the original pose is captured and reproduced. Similarly, cartoon-based anonymization methods replicate signing with animated avatars but often miss key details like facial expressions and hand configurations \citep{Tze_2022}.

\citet{battisti-etal-2024-person} found that pose estimation alone does not conceal signer identity. They noted signers could still be recognized from pose data, highlighting the need for advanced anonymization techniques to better protect privacy. Our work addresses this gap by proposing an appearance transfer to help obfuscate sign language poses.

\section{Method}\label{sec:method}

Our appearance transfer approach focuses on altering the appearance of the signer in a pose sequence while preserving the underlying sign information. The method assumes that the video starts from a relaxed posture, not mid-signing.

Given a \textcolor{blue}{pose sequence by signer $\alpha$ ($P_\alpha$)}, and a \textcolor{orange}{single pose frame by signer $\beta$ ($P_\beta$)}, both poses are normalized to a common scale based on shoulder width, using the \texttt{pose-format} \cite{moryossef2021pose-format} library. 
The appearance of both signers is assumed to exist in the first frame of each pose.

Ignoring the hands, to transfer the appearance of signer $\beta$ to the video by signer $\alpha$, we modify the pose sequence by removing the appearance of $\alpha$ and adding the appearance of $\beta$ (Equation \ref{eq:anon}).

\begin{equation}\label{eq:anon}
    \hat{P_\alpha} = \textcolor{blue}{P_\alpha} - \textcolor{blue}{P_\alpha^{\textcolor{black}{0}}} + \textcolor{orange} {P_\beta^{\textcolor{black}{0}}}
\end{equation}

To perform a standardized anonymization, we choose person $\beta$ as the mean frame in a large sign language dataset (Figure \ref{fig:mean}).
This results in an average proportioned human, which does not specifically look similar to any individual person. We note that from an information-theoretic perspective, this approach does not guarantee anonymity.
Usage is depicted in Algorithm \ref{alg:usage}.

\begin{algorithm}
\caption{`Anonymizing' a pose sequence}
\label{alg:usage}
\begin{lstlisting}[language=Python]
  from pose_format import Pose
  from pose_anonymization.appearance \
      import remove_appearance

  with open("example.pose", "rb") as f:
      pose = Pose.read(f.read())

  pose = remove_appearance(pose)
\end{lstlisting}
\end{algorithm}

\section{Qualitative Evaluation}\label{sec:qualitative_evaluation}

This simple approach yields outstanding results. To start, we show a few pose frames from different poses, when transferred to the mean appearance (anonymized) and when transferred to the appearance of a different person (Table \ref{table:example}).

We consider a recent paper on sign language stitching and rendering \cite{moryossef2023baseline}. This paper translates spoken language text to sign language videos by identifying relevant signs from a lexicon, stitching them together in a smart way (cropping neutral positions and smoothing the transition), and then rendering a video using a rendering model, trained on a single interpreter. We introduce a single intervention---after finding relevant lexicon items, we transfer the appearance of the pose to be the pose of the interpreter the renderer was trained on. 

\begin{figure}[h]
    \centering
    \begin{subfigure}{0.46\linewidth}
        \centering
        \includegraphics[width=\textwidth]{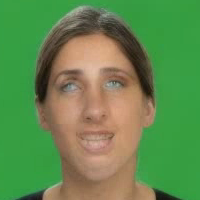}
        \caption{Without transfer}
    \end{subfigure}
    \begin{subfigure}{0.46\linewidth}
        \centering
        \includegraphics[width=\textwidth]{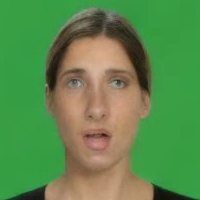}
        \caption{With transfer}
    \end{subfigure}
    \caption{Faces from ControlNet Rendering}
    \label{fig:rendering-comparison}
\end{figure}

\paragraph{Rendering} The rendering model is a Stable Diffusion model \cite{rombach2021highresolution} fine-tuned using ControlNet \cite{pose-to-image:zhang2023adding} for controllability from poses.
Since the model was trained on the appearance of a single person, it is not robust to various appearances as an input. Generally, it is not a great model, and we would like to maximize the results we get from it. Figure \ref{fig:rendering-comparison} demonstrates the rendering of the face of the original vs.\ the new pose. We can see that when transferring to the appearance of the interpreter the model was trained on, the results are more `human'.

\begin{figure*}[h]
    \centering    \includegraphics[width=\textwidth]{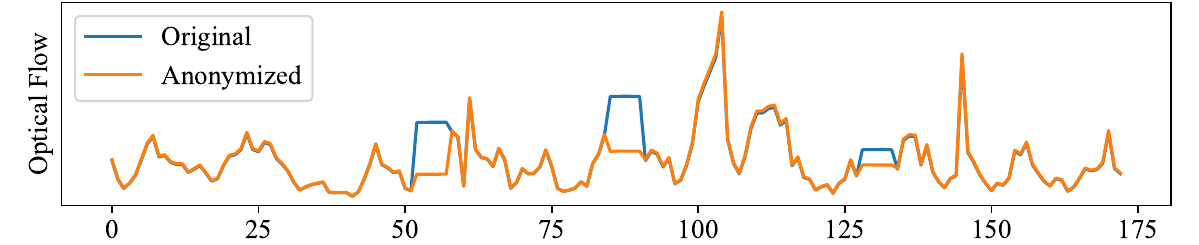}
    \caption{Optical flow (the magnitude of change between two frames) for a stitched video from four original videos and anonymized videos. Higher values represent a larger local change, and a higher area under the curve represents a larger change overall. The flow is exactly the same for all frames except for the stitching zones.}
    \label{fig:smoothing-optical-flow}
\end{figure*}
\begin{table*}
    \centering
    \begin{tabular}{lccc}
        \toprule
        Sign & Original & Anonymized & Transferred \\
        \midrule
        \makecell{Kleine\\(`small')} & \includegraphics[width=0.27\textwidth]{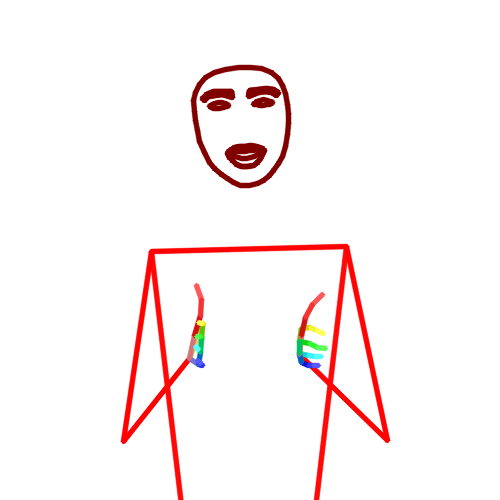} & \includegraphics[width=0.27\textwidth]{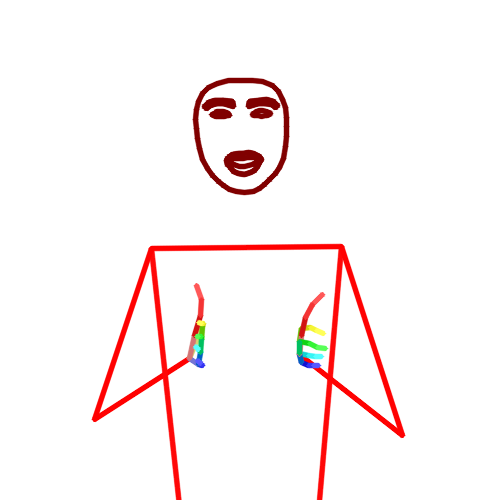} & \includegraphics[width=0.27\textwidth]{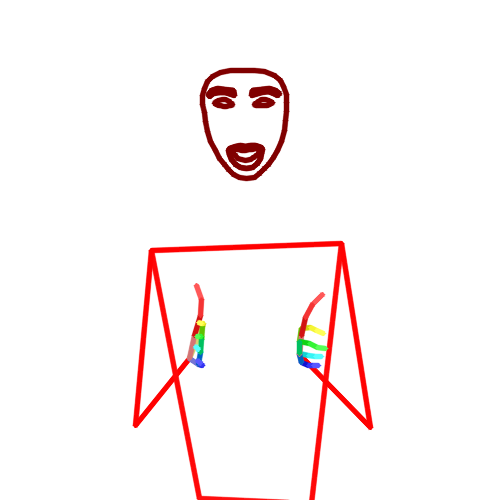} \\
        \makecell{Kinder\\(`children')} & \includegraphics[width=0.27\textwidth]{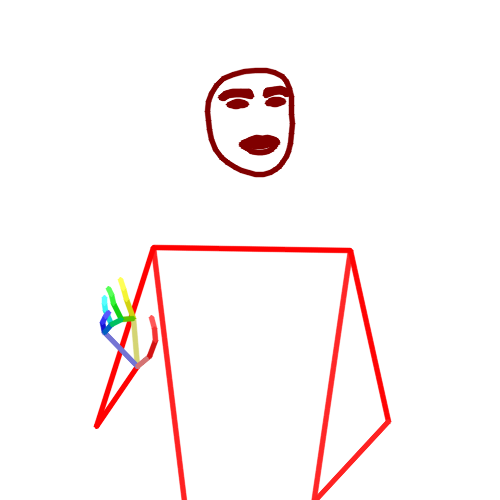} & \includegraphics[width=0.27\textwidth]{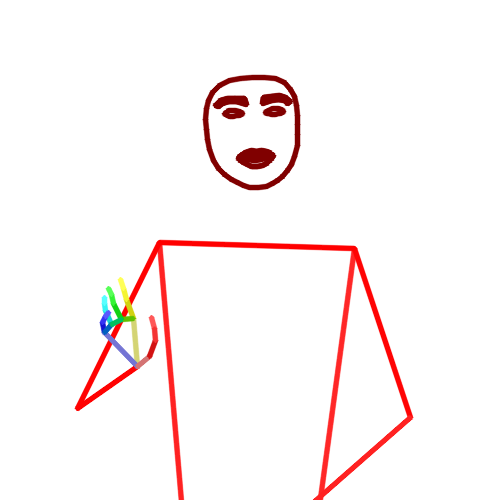} & \includegraphics[width=0.27\textwidth]{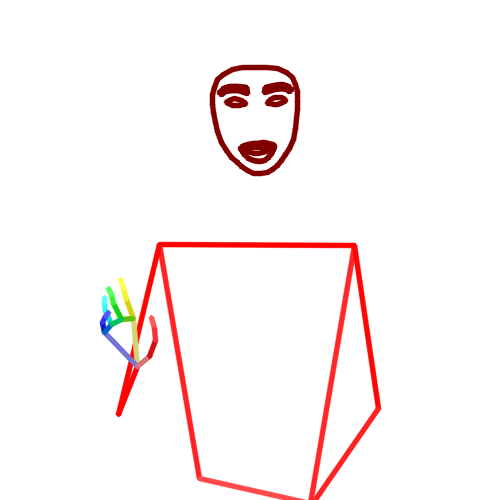} \\
        \makecell{essen\\(`eat')} & \includegraphics[width=0.27\textwidth]{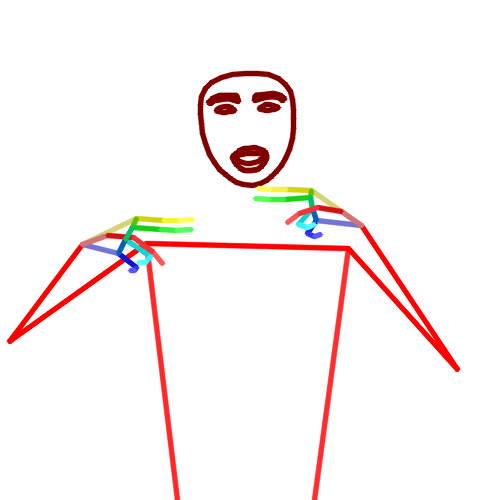} & \includegraphics[width=0.27\textwidth]{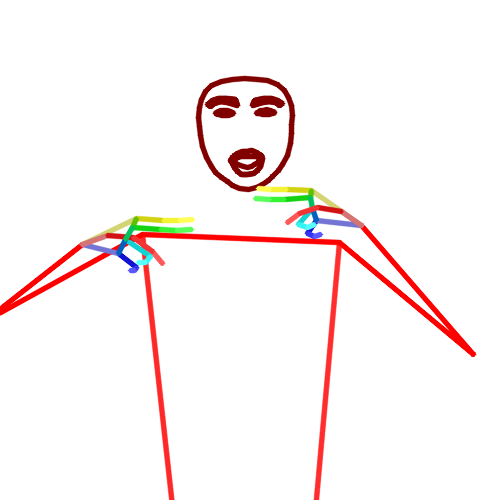} & \includegraphics[width=0.27\textwidth]{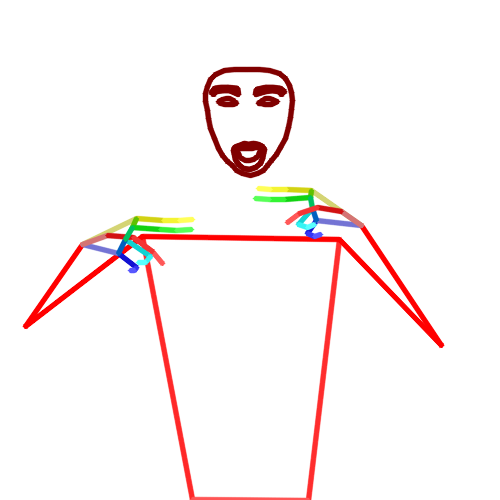} \\
        \makecell{Pizza\\(`pizza')} & \includegraphics[width=0.27\textwidth]{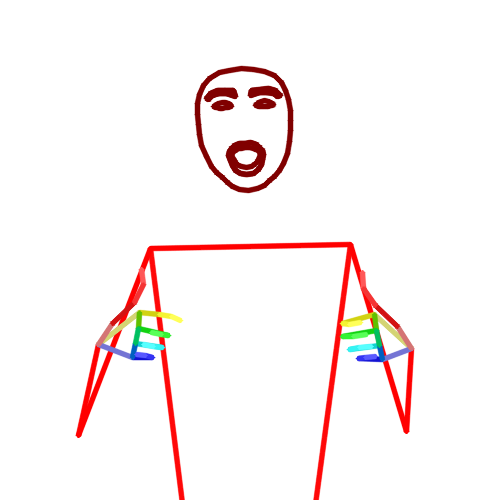} & \includegraphics[width=0.27\textwidth]{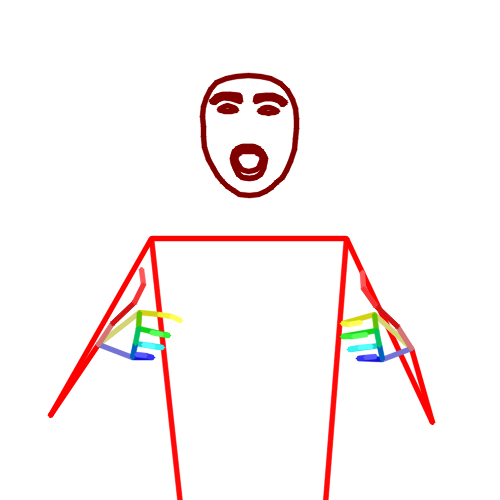} & \includegraphics[width=0.27\textwidth]{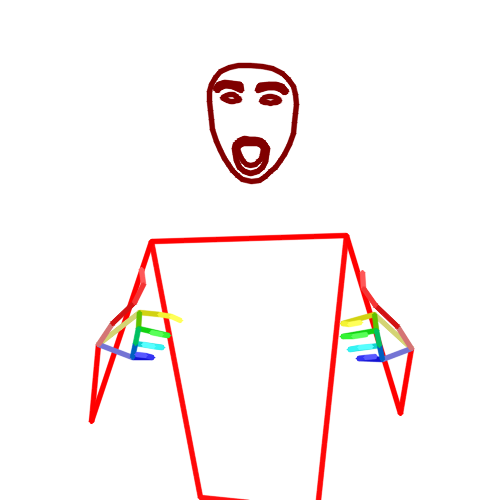} \\
        \bottomrule
    \end{tabular}
    \caption{Example of four signs. On the left, we show the middle frame from the original sign. In the middle, an anonymized version using an average pose from a large sign language dataset. On the right, appearance is transferred to be of a specific interpreter. For a video comparison, check out \url{https://github.com/sign-language-processing/pose-anonymization}.}
    \label{table:example}
\end{table*}

\paragraph{Sign Stitching} Given a uniform appearance, the stitched pose sequence is now more coherent and less jumpy. The size of different body parts does not change during the sentence, and the stitching points look smoother. When tracking optical flow across the pose sequence (Figure \ref{fig:smoothing-optical-flow}), sign transitions are smoother and less noticeable, when comparing the use of anonymized and original poses.

\section{Experiments and Results}\label{sec:experiments}

To quantify the effect of our appearance transfer method on sign language recognition, we used the code provided by \citet{moryossef2021evaluating} for both sign and signer recognition tasks. We hypothesized that transferred poses could serve as an effective data augmentation technique, allowing us to train models to a similar quality while obfuscating signer identities during both training and testing phases.

For our experiments, we used the AUTSL dataset \citep{autsl}, which includes 226 distinct lexical sign classes. Importantly, the appearance transfer process did not modify hand pose features, focusing instead on the body and face.

We trained the model under four conditions: (1) using the original pose sequences; (2) applying a single appearance transfer to the average pose shown in Figure \ref{fig:mean}; (3) transferring multiple appearances for each sample; and (4) combining all these data sources, with $10\%$ original poses, $10\%$ average poses, and $80\%$ transferred appearances. During testing, each model was evaluated on the original pose sequences, transferred to the average pose, and transferred to 10 distinct appearances, with the latter utilizing majority voting, referred to as the \emph{Transferred} method.

As shown in Table \ref{tab:sign_language_recognition_results}, no configuration outperformed the model trained and tested with the original pose sequences (top-left). However, training on a combination of original and transferred poses made the model more robust in inference on appearance-augmented data (bottom-right).

\begin{table}[h]
\centering
\small
\renewcommand{\arraystretch}{1.5} 
\resizebox{\columnwidth}{!}{%
\begin{tabular}{lccc} 
\hline
\multirow{2}{*}{\textbf{Train}} & \multicolumn{3}{c}{\textbf{Test}} \\ 
& \textbf{Original} & \textbf{Anonymized} & \textbf{Transferred} \\ \hline
(1) Original Poses              & $\textbf{80.97}\%$ & $65.82\%$ & $71.46\%$ \\
(2) Anonymized Poses       & $63.26\%$ & $64.48\%$ & $51.50\%$ \\
(3) Transferred Poses        & $67.08\%$ & $66.54\%$ & $57.32\%$ \\
(4) Combined               & $79.96\%$ & $60.88\%$ & \textbf{$76.78\%$} \\ \hline
\end{tabular}
}
\caption{Sign recognition accuracy on the AUTSL test set. `Transferred' is an ensemble of predictions from the same 10 different appearances selected randomly.}
\label{tab:sign_language_recognition_results}
\end{table}

To evaluate the extent to which our appearance transfer method obfuscates signer identity, we retrained the model using the original pose sequences but replaced the final sign classification layer with a signer classification layer, freezing the rest of the network as per \citet{sant23_interspeech}. 

When trained and tested on the original poses, the model achieved $80.18\%$ accuracy in identifying the signer, demonstrating the existence of identifiable traits. When trained and tested on anonymized poses, accuracy dropped to $65.34\%$, and with transferred poses, it fell further to $52.20\%$. These results indicate that while our method significantly reduces identifiable information, it does not eliminate it, as random chance would yield only $3.23\%$ accuracy.

\section{Conclusions}\label{sec:conclusion}

We presented a method for appearance transfer in sign language poses, allowing the alteration of a signer's appearance within a pose sequence while preserving essential signing information. By normalizing poses and selectively transferring appearance from another individual—excluding hand geometry to maintain natural movement—we achieved smooth and coherent results in sign rendering and stitching tasks.

Our qualitative evaluation shows that the appearance transfer effectively smooths pose transitions and enhances the visual coherence of stitched sign sequences. However, the quantitative results indicate that while the method helps anonymize signer identity, it can negatively impact sign language recognition performance.

\section*{Limitations}

We believe that the balance between privacy and utility is to remove all information except for the choice of signs. This is similar to how spoken language text makes speech anonymous to the degree of word choice. Practically, for anonymizing sign language videos, we propose the combination of sign language segmentation \cite{segmentation:moryossef-etal-2023-linguistically} with phonological sign language transcription. The bottleneck that transcribed sign segments introduce guarantees the removal of identifying information such as appearance, prosodic cues, and movement patterns. Then, a sign language synthesis component should synthesize the transcribed signing sequence back into video.

One major limitation of our study is the lack of human evaluation. While the method aims to preserve essential signing information, it's crucial to assess whether altering the signer's appearance affects the naturalness and comprehensibility of the signs for human viewers, especially in real-world contexts. Evaluating whether the anonymized or transferred appearances still allow viewers to recognize or identify individual signers is key to ensuring the method's success in obfuscating identity. This evaluation will provide insight into how well the technique balances privacy with the utility and intelligibility of the sign content.

\section*{Acknowledgements}

This work was funded by the SIGMA project (G-95017-01-07) at the Digital Society Initiative (DSI), University of Zurich, and by \url{sign.mt} ltd.

\bibliography{background,anthology,custom}

\begin{thebibliography}{17}
\expandafter\ifx\csname natexlab\endcsname\relax\def\natexlab#1{#1}\fi

\bibitem[{Battisti et~al.(2024)Battisti, van~den Bold, G{\"o}hring, Holzknecht, and Ebling}]{battisti-etal-2024-person}
Alessia Battisti, Emma van~den Bold, Anne G{\"o}hring, Franz Holzknecht, and Sarah Ebling. 2024.
\newblock \href {https://aclanthology.org/2024.signlang-1.2} {Person identification from pose estimates in sign language}.
\newblock In \emph{Proceedings of the LREC-COLING 2024 11th Workshop on the Representation and Processing of Sign Languages: Evaluation of Sign Language Resources}, pages 13--25, Torino, Italia. ELRA and ICCL.

\bibitem[{Bragg et~al.(2020)Bragg, Koller, Caselli, and Thies}]{bragg2020}
Danielle Bragg, Oscar Koller, Naomi Caselli, and William Thies. 2020.
\newblock \href {https://doi.org/10.1145/3373625.3417024} {Exploring collection of sign language datasets: Privacy, participation, and model performance}.
\newblock In \emph{Proceedings of the 22nd International ACM SIGACCESS Conference on Computers and Accessibility}, ASSETS '20, New York, NY, USA. Association for Computing Machinery.

\bibitem[{{European Parliament} and {Council of the European Union}(2016)}]{GDPR2016a}
{European Parliament} and {Council of the European Union}. 2016.
\newblock \href {https://data.europa.eu/eli/reg/2016/679/oj} {Regulation ({EU}) 2016/679 of the {European} {Parliament} and of the {Council}}.

\bibitem[{Goodfellow et~al.(2014)Goodfellow, Pouget-Abadie, Mirza, Xu, Warde-Farley, Ozair, Courville, and Bengio}]{goodfellow2014generative}
Ian Goodfellow, Jean Pouget-Abadie, Mehdi Mirza, Bing Xu, David Warde-Farley, Sherjil Ozair, Aaron Courville, and Yoshua Bengio. 2014.
\newblock \href {https://proceedings.neurips.cc/paper_files/paper/2014/file/5ca3e9b122f61f8f06494c97b1afccf3-Paper.pdf} {Generative adversarial nets}.
\newblock In \emph{Advances in Neural Information Processing Systems}, volume~27. Curran Associates, Inc.

\bibitem[{Hanke et~al.(2020)Hanke, Schulder, Konrad, and Jahn}]{dataset:hanke-etal-2020-extending}
Thomas Hanke, Marc Schulder, Reiner Konrad, and Elena Jahn. 2020.
\newblock \href {https://www.aclweb.org/anthology/2020.signlang-1.12} {Extending the {P}ublic {DGS} {C}orpus in size and depth}.
\newblock In \emph{Proceedings of the LREC2020 9th Workshop on the Representation and Processing of Sign Languages: Sign Language Resources in the Service of the Language Community, Technological Challenges and Application Perspectives}, pages 75--82, Marseille, France. European Language Resources Association (ELRA).

\bibitem[{Isard(2020)}]{Isard2020ApproachesTT}
Amy Isard. 2020.
\newblock \href {https://api.semanticscholar.org/CorpusID:219306343} {Approaches to the anonymisation of sign language corpora}.
\newblock In \emph{SIGNLANG}.

\bibitem[{Moryossef et~al.(2023{\natexlab{a}})Moryossef, Jiang, M{\"u}ller, Ebling, and Goldberg}]{segmentation:moryossef-etal-2023-linguistically}
Amit Moryossef, Zifan Jiang, Mathias M{\"u}ller, Sarah Ebling, and Yoav Goldberg. 2023{\natexlab{a}}.
\newblock \href {https://doi.org/10.18653/v1/2023.findings-emnlp.846} {Linguistically motivated sign language segmentation}.
\newblock In \emph{Findings of the Association for Computational Linguistics: EMNLP 2023}, pages 12703--12724, Singapore. Association for Computational Linguistics.

\bibitem[{Moryossef et~al.(2021{\natexlab{a}})Moryossef, M\"{u}ller, and Fahrni}]{moryossef2021pose-format}
Amit Moryossef, Mathias M\"{u}ller, and Rebecka Fahrni. 2021{\natexlab{a}}.
\newblock pose-format: Library for viewing, augmenting, and handling .pose files.
\newblock \url{https://github.com/sign-language-processing/pose}.

\bibitem[{Moryossef et~al.(2023{\natexlab{b}})Moryossef, M{\"u}ller, G{\"o}hring, Jiang, Goldberg, and Ebling}]{moryossef2023baseline}
Amit Moryossef, Mathias M{\"u}ller, Anne G{\"o}hring, Zifan Jiang, Yoav Goldberg, and Sarah Ebling. 2023{\natexlab{b}}.
\newblock \href {https://github.com/ZurichNLP/spoken-to-signed-translation} {An open-source gloss-based baseline for spoken to signed language translation}.
\newblock In \emph{2nd International Workshop on Automatic Translation for Signed and Spoken Languages (AT4SSL)}.
\newblock Available at: \url{https://arxiv.org/abs/2305.17714}.

\bibitem[{Moryossef et~al.(2021{\natexlab{b}})Moryossef, Tsochantaridis, Dinn, Camgöz, Bowden, Jiang, Rios, Müller, and Ebling}]{moryossef2021evaluating}
Amit Moryossef, Ioannis Tsochantaridis, Joe Dinn, Necati~Cihan Camgöz, Richard Bowden, Tao Jiang, Annette Rios, Mathias Müller, and Sarah Ebling. 2021{\natexlab{b}}.
\newblock \href {https://doi.org/10.1109/CVPRW53098.2021.00382} {Evaluating the immediate applicability of pose estimation for sign language recognition}.
\newblock In \emph{2021 IEEE/CVF Conference on Computer Vision and Pattern Recognition Workshops (CVPRW)}, pages 3429--3435.

\bibitem[{Rombach et~al.(2021)Rombach, Blattmann, Lorenz, Esser, and Ommer}]{rombach2021highresolution}
Robin Rombach, Andreas Blattmann, Dominik Lorenz, Patrick Esser, and Björn Ommer. 2021.
\newblock \href {http://arxiv.org/abs/2112.10752} {High-resolution image synthesis with latent diffusion models}.

\bibitem[{Rust et~al.(2024)Rust, Shi, Wang, Camgoz, and Maillard}]{Rust2024TowardsPS}
Phillip Rust, Bowen Shi, Skyler Wang, Necati~Cihan Camgoz, and Jean Maillard. 2024.
\newblock \href {https://api.semanticscholar.org/CorpusID:267681849} {Towards privacy-aware sign language translation at scale}.
\newblock In \emph{Annual Meeting of the Association for Computational Linguistics}.

\bibitem[{Sant and Escolano(2023)}]{sant23_interspeech}
Gerard Sant and Carlos Escolano. 2023.
\newblock \href {https://doi.org/10.21437/Interspeech.2023-2050} {Analysis of acoustic information in end-to-end spoken language translation}.
\newblock In \emph{INTERSPEECH 2023}, pages 52--56.

\bibitem[{Saunders et~al.(2021)Saunders, Camg{\"o}z, and Bowden}]{anonysign}
Ben Saunders, Necati~Cihan Camg{\"o}z, and Richard Bowden. 2021.
\newblock \href {https://doi.org/10.1109/FG52635.2021.9666984} {Anonysign: Novel human appearance synthesis for sign language video anonymisation}.
\newblock In \emph{2021 16th IEEE International Conference on Automatic Face and Gesture Recognition (FG 2021)}, pages 1--8.

\bibitem[{Sincan and Keles(2020)}]{autsl}
Ozge~Mercanoglu Sincan and Hacer~Yalim Keles. 2020.
\newblock \href {https://doi.org/10.1109/ACCESS.2020.3028072} {Autsl: A large scale multi-modal turkish sign language dataset and baseline methods}.
\newblock \emph{IEEE Access}, 8:181340--181355.

\bibitem[{Tze et~al.(2022)Tze, Filntisis, Roussos, and Maragos}]{Tze_2022}
Christina~O. Tze, Panagiotis~P. Filntisis, Anastasios Roussos, and Petros Maragos. 2022.
\newblock \href {https://doi.org/10.1109/IVMSP54334.2022.9816293} {Cartoonized anonymization of sign language videos}.
\newblock In \emph{2022 IEEE 14th Image, Video, and Multidimensional Signal Processing Workshop (IVMSP)}, pages 1--5.

\bibitem[{Zhang and Agrawala(2023)}]{pose-to-image:zhang2023adding}
Lvmin Zhang and Maneesh Agrawala. 2023.
\newblock \href {http://arxiv.org/abs/2302.05543} {Adding conditional control to text-to-image diffusion models}.

\end{thebibliography}
\bibliographystyle{acl_natbib}

\end{document}